\documentclass[letterpaper, 10 pt, conference]{ieeeconf}  

\IEEEoverridecommandlockouts                              

\makeatletter
\let\NAT@parse\undefined
\makeatother

\usepackage{graphics} 
\usepackage{graphicx}
\usepackage{grffile} 
\usepackage{amsmath} 
\usepackage{amssymb}  
\usepackage{color}
\usepackage{bm}
\usepackage{url}

\usepackage[noend]{algpseudocode}
\usepackage{algorithm}
\usepackage{makecell}
\usepackage[square, comma, numbers,sort&compress]{natbib}
\usepackage{multirow}
\let\labelindent\relax
\usepackage{enumitem}
\usepackage[table,dvipsnames]{xcolor}
\usepackage[pagebackref=true,breaklinks=true,colorlinks,bookmarks=false]{hyperref}
\usepackage{times}
\usepackage{textcomp}
\usepackage{booktabs}
\usepackage{threeparttable} 
\usepackage{siunitx}

\usepackage{float}

\usepackage{dblfloatfix}

\usepackage{multicol}

\usepackage{wrapfig}

\hypersetup{
linkcolor=BrickRed
,citecolor=Green
,filecolor=Mulberry
,urlcolor=NavyBlue
,menucolor=BrickRed
,runcolor=Mulberry
,linkbordercolor=BrickRed
,citebordercolor=Green
,filebordercolor=Mulberry
,urlbordercolor=NavyBlue
,menubordercolor=BrickRed
,runbordercolor=Mulberry
}

\newcommand\blfootnote[1]{%
  \begingroup
  \renewcommand\thefootnote{}\footnote{#1}%
  \addtocounter{footnote}{-1}%
  \endgroup
}

\title{\LARGE \bf
Deep Weakly Supervised Positioning
}

\author{Ruoyu Wang$^{1*}$, XuchuXu$^{1*}$, Li Ding$^{2}$, Yang Huang$^{1}$, Chen Feng$^{1\dagger}$
\\ \url{https://ai4ce.github.io/DeepGPS/}}

\begin{document}
\maketitle
\begin{abstract}
PoseNet can map a photo to the position where it is taken, which is appealing in robotics. However training PoseNet requires full supervision, where ground truth positions are non-trivial to obtain. Can we train PoseNet \textit{without knowing the ground truth positions} for each observation? We show that \textit{this is possible via constraint-based weak-supervision}, leading to the proposed framework: DeepGPS. Particularly, using wheel-encoder-estimated distances traveled by a robot along random straight line segments as constraints between PoseNet outputs, DeepGPS can achieve a relative positioning error of less than 2\%. Moreover, training DeepGPS can be done as auto-calibration with almost no human attendance, which is more attractive than its competing methods that typically require careful and expert-level manual calibration. We conduct various experiments on simulated and real datasets to demonstrate the general applicability, effectiveness, and accuracy of DeepGPS, and perform a comprehensive analysis of its robustness. 
\vspace{-11mm}
\end{abstract}
\blfootnote{$^1$~New York University, Brooklyn, NY 11201.
\texttt{\{ruoyuwang, xuchuxu,cfeng\}@nyu.edu}}

\blfootnote{$^2$~University of Rochester, Rochester, NY 14627. \texttt{l.ding@rochester.edu}}

\blfootnote{$^*$ Equal contributions.}

\blfootnote{$^{\dagger}$~The corresponding author is Chen Feng. \texttt{ cfeng@nyu.edu}}

\section{Introduction}
\label{sec:intro}



Visual localization/positioning has been a fundamental problem in robotics and computer vision which draws attention from researchers for a long time. Given a local observation, the goal of visual localization is to recover the sensor position in the scene where the local observation is captured. Depending on the modality of local observations, visual localization can be based on either images or point clouds. The ability of localizing the sensor plays a vital role in a variety of applications including virtual and augmented reality (VR/AR)~\cite{Newcombe:DTAM:ICCV11}, 3D reconstruction~\cite{Tanskanen:LiveMetricReconMobile:ICCV13}, and robotics~\cite{Lim:RTMono6DOF:IJRR15}. In robotics particularly, commercial visual localization systems like Vicon are expensive yet popular options. Conventional visual localization approaches~\cite{Donoser:DiscrimFeat2PtMatch:CVPR14,Liu:Glb2D3DMatch:ICCV17,Ding:FuseSfMLidar:ICASSP17} typically rely on finding corresponding feature points/lines/planes~\cite{taguchi2013point,pumarola2017pl} between the local observation and the global scene, which are then used to determine the sensor position, typically within the RANSAC~\cite{Fischler:RANSAC:CACM81} framework to handle mismatched features.

Recently, deep learning methods have demonstrated compelling improvements in geometric computer vision problems, and many efforts have been made for visual localization from images~\cite{Sarlin:Coare2FineCamLoc:CVPR19,Avraham:EMPNet:ICCV19} and point clouds~\cite{Dube:SegMap:RSS18,Wang:DeepPCO:IROS19}. Among them, PoseNet~\cite{Kendall:PoseNet:ICCV2015} is one of the first that formulates the visual localization as the absolute pose regression problem where the global scene is implicitly represented inside a deep neural network (DNN). Since then, this approach has been studied extensively. Despite different network architectures or loss functions, these methods follow the same supervised learning pipeline where DNNs are first trained on a large set of observations along with the corresponding ground truth sensor poses and then directly used to predict sensor poses for unseen observations. The performance of this approach strongly relies on the quantity and the quality of available training data. Benchmarking results~\cite{Sattler:LimitationOfAPR:CVPR19} have shown that the performance would significantly drop when training data is limited. However, large-scale high-quality ground truth positions for input observations are non-trivial to obtain. The ground truth data collection requires either external devices such as the Vicon system or high-end GPS, or complex 3D reconstruction methods such as Structure-from-Motion (SfM) or Simultaneous Localization and Mapping (SLAM) that often calls for expert efforts. While a trained PoseNet can provide fast and efficient pose estimation, the requirement of ground truth data makes it less convenient for roboticists.

\begin{figure}[t]
    \hspace{1mm}
    \begin{minipage}[b]{0.45\linewidth}
        \centering
        \includegraphics[width=1\linewidth]{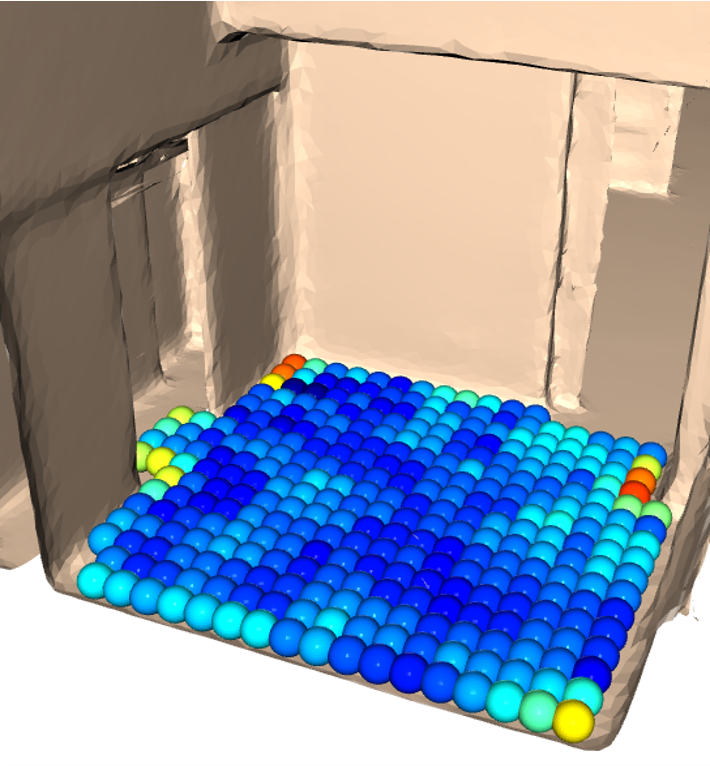}
    \end{minipage}
    \hspace{2mm}
    \begin{minipage}[b]{0.45\linewidth}
        \centering
        \includegraphics[  width=1\linewidth]{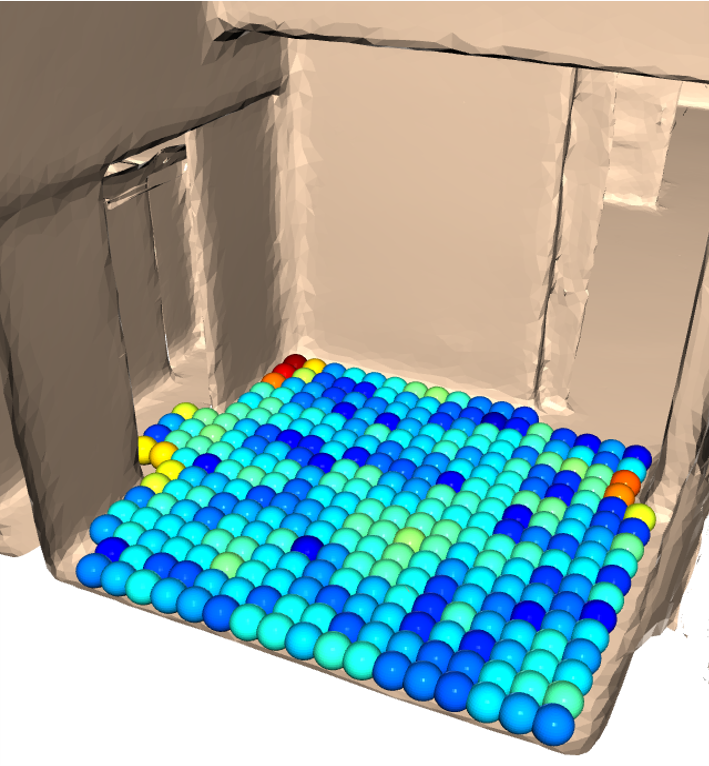}
    \end{minipage}
    \caption{Visual localization by the proposed weakly-supervised DeepGPS (left) and the fully-supervised PoseNet~\cite{Kendall:PoseNet:ICCV2015} (right) on an indoor environment from the Gibson dataset~\cite{xiazamirhe2018gibsonenv}. Each sphere shows the testing location and the color-coded error (the darker the better).  Best viewed in color.\label{fig:3d_comp_deepgps_posenet}}
\end{figure}

In this paper, we explore a different approach, DeepGPS, for visual localization using DNNs that does not require the absolute sensor positions for each observation, yet still achieves a comparable performance as PoseNet (see Figure~\ref{fig:3d_comp_deepgps_posenet}). We notice that a mobile robot can often easily travel along a random straight line and accurately estimate the traveled distance using odometry from wheel encoders. This leads to \textit{the key idea behind the proposed DeepGPS}: instead of supervised learning on sensor positions for visual localization (such as PoseNet), we adopt the constraint-based weakly supervised learning, where the constraints come from the distances between two sensor positions where the local observations are captured. The main advantage of this weakly supervised approach is that the large-scale data collection can be performed fully automatically with almost no human attendance. The training can be considered as an auto-calibration, which is more attractive than alternative approaches, especially for indoor or lab settings, which typically involve expert-level manual calibration. To summarize, we make the following contributions:
\begin{itemize}
	\item We find that it is feasible to train DNNs for visual localization via constraint-based weakly supervised learning without ground truth positions.
	\item We propose DeepGPS framework that uses distances between sensors, rather than sensor position itself, as the indirect supervision signals.
	\item We show by comprehensive experiments that DeepGPS achieves less than 2\% relative error with various network architectures for different sensing modalities.
	\item We perform robustness analysis of DeepGPS to improve our understanding on its training data requirements.
\end{itemize} 

\section{Related Works}
\label{sec:related_works}

There has been a large body of work on positioning system using various sensing modalities. We refer readers to~\cite{brena2017evolution,zafari2019survey} for comprehensive surveys of non-vision-based methods and focus our discussion on the vision-based methods.


\noindent\textbf{Visual localization.} Prior works on visual localization generally fall into two groups. The methods in the first category are based on matching features between the local and the global scene. The scene environment can be explicitly represented as a 3D model, which is reconstructed from images or directly captured by 3D scanners. The correspondences can be established by matching hand-crafted~\cite{Lowe:SIFT:IJCV04,Rublee:ORB:ICCV11} or learning-based~\cite{Dube:SegMatch:ICRA17,Dube:SegMap:RSS18} features. The downside is that these approaches require an explicit representation of the 3D model or have to store all descriptors and camera poses in memory. Additionally, several works use machine learning and deep learning methods to directly regress 3D coordinates of pixels in the image~\cite{Brachmann:LearnLessMore6DCamLoc:CVPR18,Massiceti:RFvsNNCamLoc:ICRA17}. While learning to predict 3D coordinates from image patches yields accurate sensor poses, these approaches may fail to handle large-scale scene~\cite{Taira:InLoc:CVPR18,Schonberger:SemanticVisLoc:CVPR18}.

Rather than learning to find corresponding points between local and global scene, another type of learning-based approach, absolute pose regression (APR), use DNNs to model the full localization pipeline. In this scenario, the global environment is implicitly represented by the weights of DNNs. The core idea is to build DNN models to directly predict the sensor positions for the input observations. The first APR-based approach is PoseNet~\cite{Kendall:PoseNet:ICCV2015} that uses a modified GoogLeNet~\cite{SzegedyGoogLeNet:CVPR15} to regress camera poses. The network is trained by minimizing the Euclidean distance between the predicted poses and the ground truth poses. Since then, the APR-based approach for visual localization has been explored extensively. For example, method in~\cite{Kendall:GeoLoss4PoseReg:CVPR2017} incorporate additional loss terms such as the reprojection error and the weighted position and rotation loss. In addition to the novel loss function, different network architectures are proposed for extracting the fine-grained features~\cite{Melekhov:CamLocHourglassNet:ICCVW17} or for handling sequences of inputs~\cite{Henriques:MapNetRNN:CVPR18}. Moreover, view synthesis techniques~\cite{Wu:DelveCNN4CamLoc:ICRA2017,Naseer:RegMonoCam6DOFOutdoor:IROS2017} are integrated into the APR-based approach to enlarge the training dataset and to pose constraints on sensor poses. Methods in~\cite{Brahmbhatt:GeoLearnMap4CamLoc:CVPR18,Radwan:VLocNetPP:RAL18} exploit other available supervision signals, including GPS, IMU, and VO, to improve the accuracy of localization. Despite various network architectures and loss functions, these methods follow the supervised pipeline and require a fully labeled training dataset that contains a massive amount of local observations and the associated ground truth positions. As noted in Section~\ref{sec:intro}, such ground truth positions are non-trivial to obtain and therefore pose a challenge for this supervised learning approach to be effectively deployed.

\noindent\textbf{Learning-based visual odometry.} Another related problem to visual localization is the visual odometry (VO) that incrementally estimates the ego-motion of the sensor using a sequence of observations. We briefly review the learning-based VO and refer readers to the survey paper~\cite{Fraundorfer:VOSurveyPart2:RAM12} for broader context on the conventional methods. DeepVO~\cite{Wang:DeepVO:ICRA2017} and VidLoc~\cite{Clark:VidLoc:CVPR2017} propose to use a combined CNN and LSTM networks for monocular VO by taking the advantage that CNN and LSTM are capable to extract features from the input images and to incorporate temporal information, respectively. In addition to monocular camera, different sensors are incorporated into learning-based method that allow the information from independent sensors to be fused in a complementary way, including lidar~\cite{Ma:SelSupLidar&cam:ICRA2019}, GPS~\cite{Pillai:TowVEGO;IROS2017}, or IMU~\cite{Wang:LearnInertia:ICASSP2019}. Several works focus on unsupervised learning for the VO problem by jointly estimating the depth maps for the input RGB images~\cite{Zhou:SfMLearner:CVPR17,Li:UnDeepVO:ICRA2018}.

\begin{figure}[t]
\vspace{3mm}
    \centering
    \includegraphics[width=0.9\linewidth]{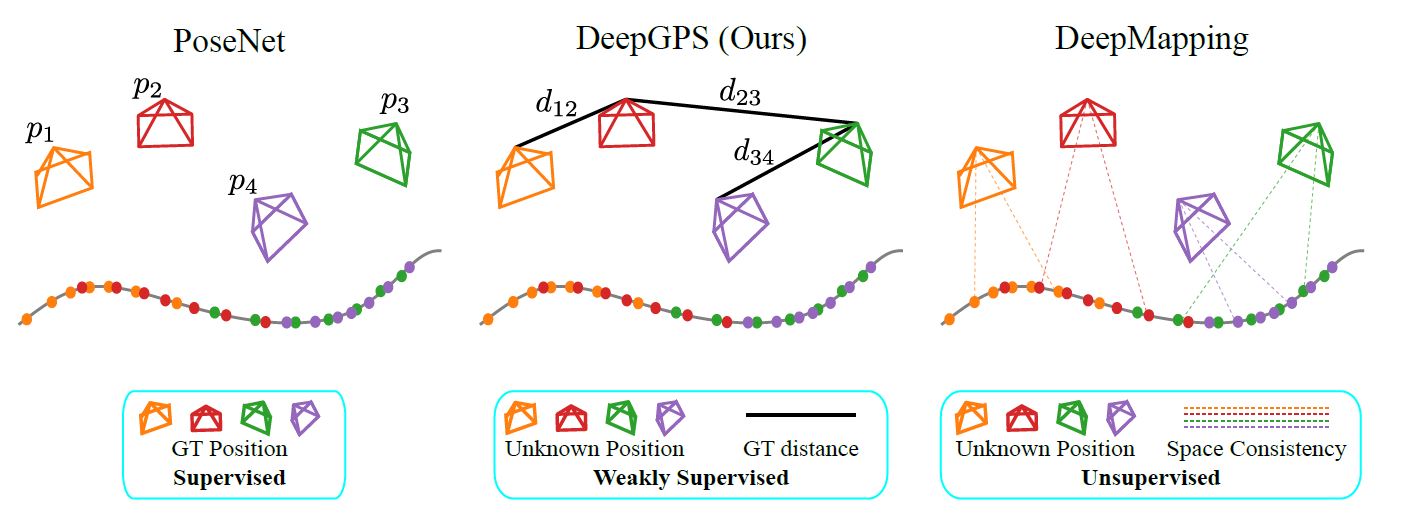}
    \caption{Comparison of training procedures for various positioning methods: PoseNet~\cite{Kendall:PoseNet:ICCV2015} (supervised), the proposed DeepGPS (weakly supervised), and DeepMapping~\cite{Ding:DeepMapping:CVPR19} (unsupervised). PoseNet requires ground truth (GT) poses. DeepMapping requires point clouds for the free-space consistency loss. Our DeepGPS only requires line segment traveling and distance measuring, which is easy for most of the mobile robots. DeepGPS can also use different observation modalities, including images and point clouds. \label{fig:compr_learning_approach}}
    \vspace*{-0.15in}
\end{figure}

\noindent\textbf{Beyond supervised learning.} Most deep learning based methods for visual localization are fully supervised. As noted, ground truth data collection is non-trivial. Therefore, a few alternative learning strategies have been investigated as well. In Fig.~\ref{fig:compr_learning_approach}, we compare several training procedure for visual localization. Unsupervised learning, for example, avoids labeled data collection. The recent DeepMapping~\cite{Ding:DeepMapping:CVPR19} introduces an unsupervised framework for multiple point clouds registration and mapping, where the supervision signals are derived from the free-space consistency among point clouds. The trained model is able to perform coarse re-localization for unseen point clouds that are in close vicinity to the registered point clouds. The main drawback of this method is that it is only applicable to point cloud modality. 

    
    


\section{DeepGPS Method}
\label{sec:method}
In this section, we first formulate the {visual positioning} problem in our weakly-supervised paradigm. Then, we introduce the different settings {of our method for different observation modalities. Finally, we introduce two automatic data collection strategies for mobile robots.} 

\subsection{Constraint-based Weak Supervision}

\newcommand{\norm}[1]{\left\lVert#1\right\rVert}
\newcommand{\lc}{\left ( }
\newcommand{\rc}{\right ) }
\newcommand{\lsq}{\left [ }
\newcommand{\rsq}{\right ] }
\newcommand{\vc}{\mathbf{c}}
\newcommand{\vp}{\mathbf{p}}
\newcommand{\vm}{\mathbf{m}}
\newcommand{\vx}{\mathbf{x}}
\newcommand{\vy}{\mathbf{y}}
\newcommand{\dnn}{f_\theta}



In the supervised learning setting, the goal of visual positioning is to find a function $f_\theta\lc\vx\rc$, modeled as a DNN with learnable parameters $\theta$, to predict sensor position $\vp$. Given a mostly static environment, $f$ is a bijection if the local observation $\vx$ is independent with sensor orientation, and otherwise a surjection. The training dataset $\mathcal{D}_{s} = \left\{ \lc \vx_i, \vp_i \rc \right\}_{i=1}^{N}$ contains a set of local observations $\vx_i$ associated with the ground truth positions $\vp_i$. The network parameters are obtained by minimizing the loss function:
\begin{equation}
\ell_{s} \lc \theta \rc = \sum_i \norm{\dnn\lc \vx_i\rc  - \vp_i},
\end{equation}
which penalizes the disagreement between network predictions and the ground truth positions.

Differently, we propose to tackle the {visual positioning} problem as constraint-based weakly supervised learning, by specifying certain constraints over the network output~\cite{Stewart:PhyConstraintSupervision:AAAI17}. 
Our training dataset is a graph $\mathcal{D}_{ws} = \lc \mathcal{V}, \mathcal{E} \rc$ where nodes $\mathcal{V} = \left\{ \vx_i \right\} _{i=1}^{N}$ are local observations, and edges $\mathcal{E} = \left\{ c_{ij} \right\}$ are 
constraints between sensor positions $\vp_i$ and $\vp_j$ defined by the constraint function $C\lc \cdot, \cdot \rc$. {Note that this graph does not need to be fully connected, i.e., $\mathcal{E}$ does not necessarily contain constraints between all pairs of $\vp_i$ and $\vp_j$. A constraint $c_{ij}$ exists in $\mathcal{E}$ only if it is accessible, i.e., when measured physically.} In general, we minimize the following loss to penalize constraint violations,
\begin{equation}\label{eq:loss_ws}
    \ell_{ws} \lc \theta \rc = \sum_{{c_{ij}}\in\mathcal{E}} \norm{ C\lc \dnn\lc\vx_i\rc, \dnn\lc\vx_j\rc \rc - c_{ij} }.
\end{equation}


As mentioned, the constraint we adopted is the Euclidean distance between two (unknown) sensor positions $i$ and $j$ where local observations are captured, i.e., $c_{ij} = \norm{\vp_i - \vp_j}_2$, and the constraint function  $C\lc \cdot, \cdot \rc$ calculates the distance between two network-predicted positions. Since wheel encoders or IMU is commonly available for mobile robots, the value of $c_{ij}$ can be measured with a high accuracy if the robot moves along the straight line between $\vp_i$ and $\vp_j$ without blocked by obstacles.

{In practice, we found that the training would converge faster if we normalized each loss term with the sum of the estimated and the ground truth distances. Thus we use the following loss function throughout our experiments:
\begin{equation}\label{eq:our-loss2}
    \ell_{ws} \lc \theta \rc = \sum_{{c_{ij}}\in\mathcal{E}} \frac{\norm{ \norm{\dnn\lc\vx_i\rc - \dnn\lc\vx_j\rc}_2 - c_{ij} }_1} { \norm{\dnn\lc\vx_i\rc - \dnn\lc\vx_j\rc}_2 + c_{ij} }.
\end{equation}}





{}

\textbf{Why it works?}
The effectiveness of DeepGPS has a connection with Multi-Dimensional Scaling (MDS), a well known technique in data mining and visualization. In effect, Eq.~\eqref{eq:loss_ws} provides a direct supervision for a function $C_\theta \lc \vx_i, \vx_j \rc$ that predicts the Euclidean distance between the positions $i$ and $j$ using the two observations, and this function has a Siamese structure denoted by $\dnn$. After the full supervision on $C_\theta$, we can use it to compute a Euclidean Distance Matrix (EDM) for a set of positions inside that environment. From MDS, the coordinates of those positions can be computed by eigen-decomposition on a centered version of EDM. If we choose this set of positions to be the densely sampled grid locations inside the environment, then in effect, we get a ``GPS sensor'' for that environment. \textit{Interestingly, DeepGPS is more appealing because we do not need to actually go through the above MDS process: the ``PoseNet'' $\dnn$ implicitly achieves it more efficiently}.


\subsection{Frameworks for different observation modalities}
The specific network architecture $f$ depends on the input modality $\vx$. We summarize the observation modalities into two main categories, observation with correspondence and observation without correspondence, and describe the network architectures accordingly. 

\subsubsection{Observation with correspondence}\label{sssec:explicit_positioning}
In certain situations, the correspondences between the observations are known. For instance, in some RF-based beacon systems, the observations are the distances to the RF signal transmitters installed in the environment. With the unique IDs decoded from the RF signals, the observations can be matched. Another example is GPS, where the observations are the distances to the particular observed satellites. The signal transmitted by each satellite can be uniquely identified. 

Without loss of generality, the situation of observation with correspondence can be described by a simple but representative model: given $M$ landmarks $\{\mathbf{m}_k\}_{k=1}^{M}$ in the environment, the observation signal $\vx_i$ can be modeled as an M-dimensional vector where each element $x^{(i)}_k$ is the distance between the landmark $\vm_k$ and the sensor location $\vp_i$. In this scenario, the positioning function $f_\theta:\mathbb{R}^M\to\mathbb{R}^D$ can be represented as a multi-layer perceptron (MLP), where $D$ is the dimension of the position coordinates.

\textbf{Relationship to explicit positioning}. 
Note that in the situations where the correspondences are known, the problem is equivalent to solving a non-linear equation system:
\begin{equation}
   \label{eqn:ba}
   \begin{cases} 
   \|\mathbf{m}_k-\mathbf{p}_i\|  =  x^{(i)}_k,\,i\in\mathbb{N}_{\leq M}\,,j\in\mathbb{N}_{\leq N} \\ 
   \|\mathbf{p}_i-\mathbf{p}_{j}\|  =  c_{jk},\,j\in\mathbb{N}_{\leq N},
   \end{cases}
\end{equation}
Once all $\mathbf{m}$ and $\mathbf{p}$ are solved, a new position $\vp'$ can be triangulated via $\mathbf{m}$ and $\vx'$. In the next section, we will compare our method with this explicit positioning approach.

\subsubsection{Observation without correspondence}
\label{sec:trans_inv}
For sensing modalities such as RGB images, correspondences between raw observations (pixels) are unknown. In this case, $f_\theta$ is usually not a bijection because $\vx$ depends on both sensor location and orientation. To achieve orientation-invariant positioning, we use  upward-viewing omnidirectional images as $\mathbf{x}$. Rotating an image taken at a position with certain orientation is equivalent to taking another image at the same location with a corresponding orientation. Then we warp the observed images $\mathbf{x}(u, v)$ from Cartesian coordinate system to polar coordinate system, i.e., $\mathbf{x}(r, \phi)$. 


\textbf{Translation-invariant CNN}. According to \cite{kayhan2020translation}, a convolutional layer is translation-invariant when the output has the same height and width as the input via circular padding. In our method, we use ResNet~\cite{he2016deep} as our backbone network and change the kernal size of the first convolution layer from 6 to 3 and replace all zero paddings with circular paddings.  

\textbf{Data augmentation}. We also randomly and circularly shift the warped image along the $\phi$-dimension and feed them into the network the during the training to simulate multiple observations at the same position with different orientations.

\subsection{Observation and ground truth distance collection.}
We adopt two strategies to collect the ground truth distance as the weak supervisions to train DeepGPS, i.e., the end-point strategy and the dense-sampling strategy.

In the end-point strategy, a robot sequentially visits random positions $\{\mathbf{p}_i\}_{i=1}^N$ in the target environment, collect local observation $\mathbf{x_i}$ at each position $\mathbf{p_i}$, and measure the ground truth distance $c_{i-1,i}$ along a straight line in between.

While the end-point strategy allows robot to freely explore the environment, this data collection strategy is inefficient because no data is collected between two consecutive points $\mathbf{p}_{i-1}$ and $\mathbf{p}_{i}$. As a result, the total traveling distance is inevitably large to collect sufficient data. To overcome this issue, we further proposes a dense-sampling strategy that is a line-segment-based data collection. Specifically, the trajectory of the robot contains $L$ random sampled line segments $\{l_k\}_{k=1}^L$. When traveling along each line segment $l_k$, observations $\{\mathbf{x}_i^{(k)}\}_{i=1}^{N_k}$ are collected at a sequence of $N_k$ points $\{\mathbf{p}_i^{(k)}\}_{i=1}^{N_k}$ along the line segments. The order of  $\mathbf{p}^{(k)}_i$ follows the movement direction of the robot. The starting point on line segment $l_k$ is $\mathbf{p}^{(k)}_1$ and $c^{(k)}_{i1}$ is recorded by the wheel encoder for all the sampling locations $i$ traveled by the robot on the $k$-th line segment. More generally,  $c^{(k)}_{ij}=|c^{(k)}_{i1}-c^{(k)}_{j1}|$ means the distance between any two sampling location $i$ and $j$ on the $k$-th line segment. Using the dense-sampling strategy, Eq.~\eqref{eq:our-loss2} becomes:
\begin{equation}\label{eq:our-loss3}
\resizebox{.9\hsize}{!}{
$
    \ell_{ws} \lc \theta \rc = \sum_{k=1}^{L}\sum_{i=1}^{N_k-1}\sum_{j=i+1}^{N_k} \frac{\norm{ \norm{\dnn(\vx^{(k)}_{i}) - \dnn(\vx^{(k)}_j)}_2 - c^{(k)}_{ij} }_1} { \norm{\dnn(\vx^{(k)}_{i}) - \dnn(\vx^{(k)}_{j})}_2 + c^{(k)}_{ij} }.
$
}
\end{equation}
The end-point strategy can be viewed as a special case of the dense-sampling strategy where $N_k=2, \forall k\in [1,L]$.

\section{Experiments}
\label{sec:experiments}
In this section, we report the performance of our method on different modalities of observations. The experiments were performed under 3 types of environments with different levels of reality: \textit{numerically simulated environment}, \textit{virtualized photo-realistic environment}, and \textit{real-world environment}. We evaluated our method for observations with correspondences in the \textit{numerically simulated environment} as a toy example, and for observations without correspondences in the \textit{virtualized photo-realistic environment} and the \textit{real-world environment} as more realistic testings. We use the Absolute Trajectory Error (ATE)~\cite{sturm2012benchmark} as the evaluation metric. We use PoseNet~\cite{Kendall:PoseNet:ICCV2015} in all the environments as a baseline for DeepGPS. For the toy example, we additionally add the explicit positioning baseline (see section~\ref{sssec:explicit_positioning}). The settings of all baseline methods are the same as DeepGPS. In our appendix, we also show experimental results on simulated 2D lidar point cloud.


\subsection{Toy example: numerically simulated environment}
\label{sec:toy_data}
\begin{figure}[t!]
\vspace{3mm}
    \centering
    \includegraphics[width=0.95\linewidth]{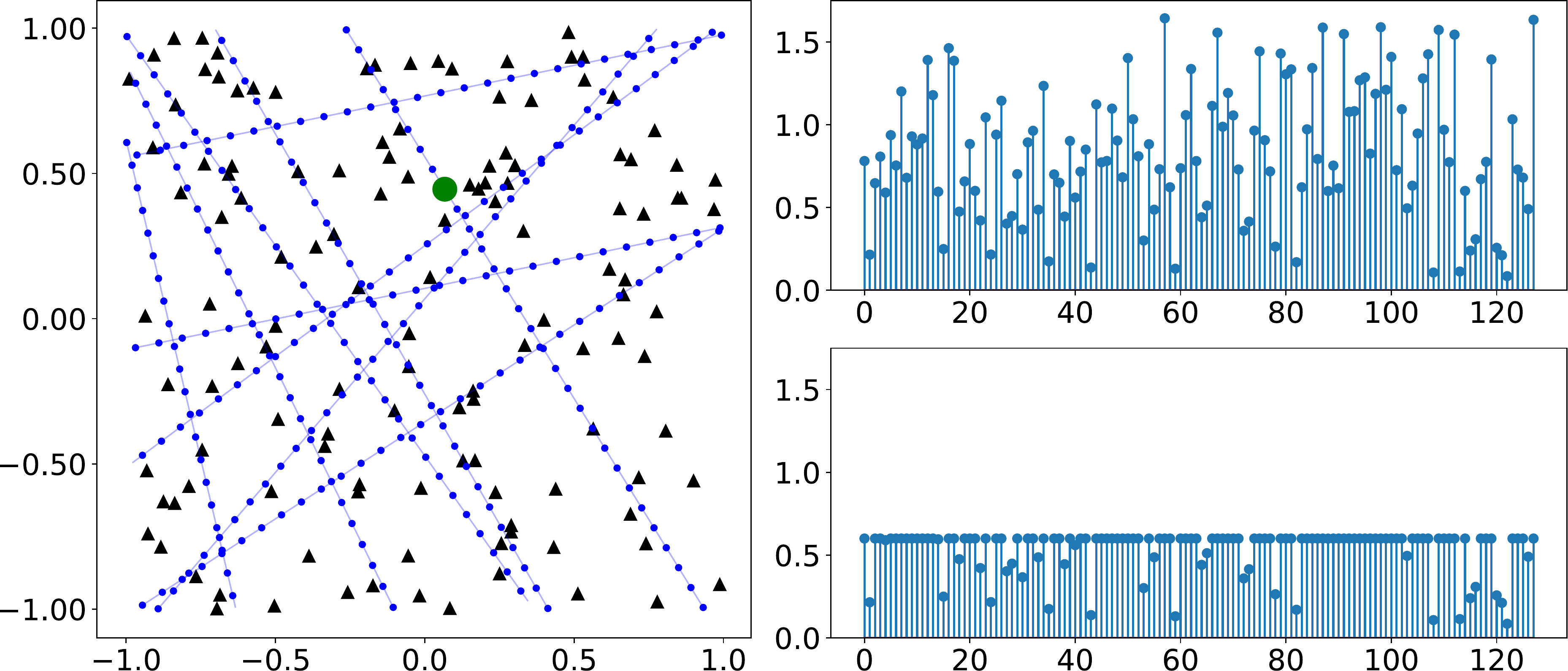}
    \caption{Data generation for the toy example experiment. The left image illustrates the sensor positions (blue points) on 10 random line segments (blue lines) and the landmarks (black triangles). The right image shows the complete (top) and the incomplete (bottom) observations captured at the green point in the left image. The observations are the distance between the sensor position and the landmarks. \label{fig:toy_traj}}
\end{figure}
\begin{figure}[t!]
    \centering
    \includegraphics[trim=0 10mm 0 0 ,width=0.48\textwidth]{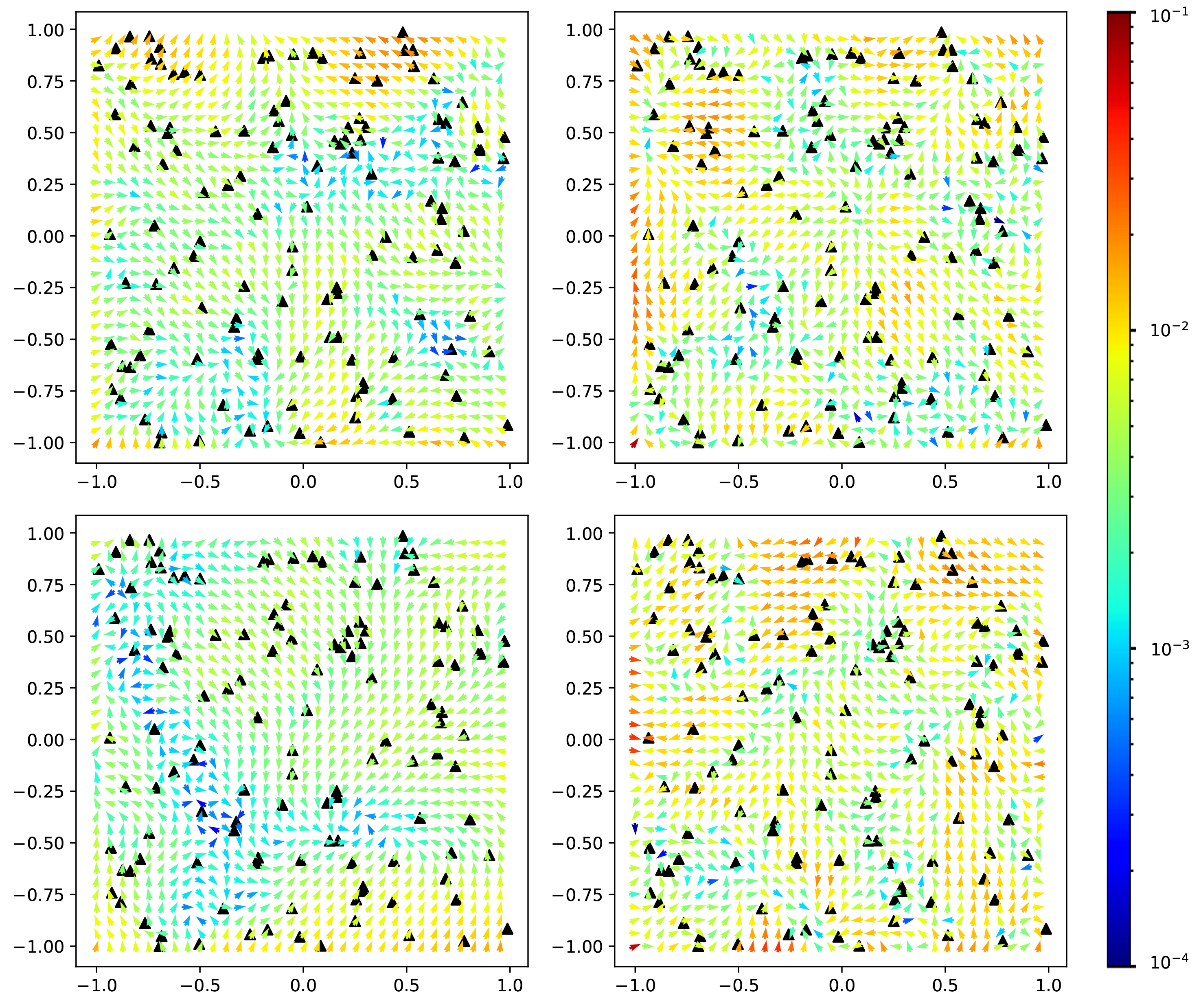}
    \caption{\textbf{Visual results of positioning error in the numerically simulated environment.} Top: PoseNet~\cite{Kendall:PoseNet:ICCV2015}, Bottom: DeepGPS. Left: complete observation, Right: incomplete observation. We show the localization errors computed from a discrete grid. The arrows indicate the error directions between the predicted positions and the ground truth, where the error magnitudes are color-coded. }
    \label{fig:toy_results}
    \vspace{-4mm}
\end{figure}
\textbf{Data generation:}
In this experiment, we use the dense-sampling strategy to collect data. The left figure in Fig.~\ref{fig:toy_traj} illustrates the data collection process. The whole environment is a 2D space bounded by $[-1, 1]$ on both dimensions. We place 128 landmarks in the environment (black triangles in Fig.~\ref{fig:toy_traj}). In the training dataset, a simulated robot starts at a random position and moves along a straight line segment (blue line in Fig.~\ref{fig:toy_traj}). Once the robot hits the boundary, it will choose another random orientation and move along a new line segment. The sampling distance is 0.02 between $\mathbf{p}_i^{(k)}$ and $\mathbf{p}_{i+1}^{(k)}$ on each line segment $l_k$. At each position, the robot measures the distance between sensor position and the landmark that forms a 128-dimensional vector. The entire training dataset contains 14,413 observations collected on 128 line segments. For testing, we uniformly sample a $128 \times 128$ grid in the environment as testing positions.


\textbf{Completeness of observations:}
In practice, the robot may not be able to observe all landmarks at a certain location. To simulate such a situation, we set a maximum observing distance $d_{max}$. If the distance between the landmark and the observing position is larger than $d_{max}$, then the distance observation to that landmark is set to $d_{max}$. The top-right and bottom-right figures in Fig.~\ref{fig:toy_traj} shows the complete and the incomplete observations, respectively. We evaluated our method with both complete observation (all landmarks can be observed) and incomplete observation (not all landmarks can be observed). We set $d_{max}=0.6$ for incomplete observation.

\textbf{Training details:}
We use an MLP with 128, 512, 512, 512, 256, 256, 128, 64, 2 neurons in each layer. The total number of training epochs is 1500 with a batch size 800. We use the Adam optimizer~\cite{Kingma:Adam:ICLR15} with learning rate 0.001 in the first 300 epochs and 0.0001 in the remaining epochs.

\begin{table}
\vspace{3mm}
\centering
\caption{Quantitative results for the toy example experiment. \label{tab:toy_quant} }
\begin{tabular}{m{1.2cm} >{\centering\arraybackslash}m{0.75cm} >{\centering\arraybackslash}m{0.7cm} >{\centering\arraybackslash}m{0.7cm} >{\centering\arraybackslash}m{0.7cm} >{\centering\arraybackslash}m{0.7cm} >{\centering\arraybackslash}m{0.7cm}}
\toprule
\multirow{2}{*}{ATE} & \multicolumn{3}{c}{Complete Observation} & \multicolumn{3}{c}{Incomplete Observation} \\ 
\cline{2-7}
& RMS & Median & Max & RMS & Median & Max \\ 
\midrule
E.P. & 0.800 & 0.754 & 1.613 & 1.131 & 0.742 & 5.048 \\ 
PoseNet\cite{Kendall:PoseNet:ICCV2015} & 0.006 & 0.004 & 0.024 & \textbf{0.007} & \textbf{0.006} & \textbf{0.069}\\ 
\cline{1-7}
DeepGPS & \textbf{0.004} & \textbf{0.004} & \textbf{0.021} & 0.009 & 0.007 & 0.111
\\
\bottomrule
\end{tabular}
 \begin{tablenotes}
      \small
      \item ``E.P.'' means explicit positioning (see section~\ref{sssec:explicit_positioning}).
    \end{tablenotes}
\vspace{-3mm}
\end{table}

\textbf{Results:}
Table~\ref{tab:toy_quant} lists the results of DeepGPS and the baseline methods. Explicit positioning does not perform well, achieving a median of 0.754 and 0.742 for the complete observation and the incomplete observation, respectively. This is because the explicit positioning method requires reasonable initialization while the learning-based method does not. Our method performs better than PoseNet~\cite{Kendall:PoseNet:ICCV2015} with complete observations. The results of incomplete observations, while slightly worse than PoseNet, still demonstrate the effectiveness of DeepGPS. It is worth noting that DeepGPS only requires weak supervisions, i.e., the distance between two observations, which are readily available in robot applications. Figure~\ref{fig:toy_results} illustrates the qualitative results. From the figure, we can see that most of the errors are less than 0.02 ($1\%$ relative to the side length of the environment). 

\begin{table*}
\vspace{3mm}
\centering
    \caption{ATE for experiments in virtualized photo-realistic environment (Gibson/Habitat-Sim). \label{tab:cnn_sim}}
\begin{tabular}{
m{1.5cm} |  
>{\centering}m{1.0cm} 
>{\centering}m{1.0cm} 
>{\centering}m{1.0cm} 
>{\centering}m{1.0cm}
>{\centering}m{1.2cm}
>{\centering}m{1cm} 
>{\centering}m{1cm}
>{\centering}m{1cm}
>{\centering}m{1cm} 
>{\centering\arraybackslash}m{1.5cm} 
}
\toprule
ATE (m) & Aloha  & Brevort & Cokeville & Delton  & Euharlee & Germfask & Islandton
& Montreal & Sodaville & \textbf{Overall} 
\\
\midrule
\multirow{3}{*}{PoseNet~\cite{Kendall:PoseNet:ICCV2015}} & \textbf{0.061} & 0.060 & \textbf{0.040} & 0.060 & \textbf{0.161} & 0.037 & 0.338 & 0.072 & 0.211 & \textbf{0.090} \\ 
& \textbf{0.037} & 0.043 & \textbf{0.032} & 0.035 & \textbf{0.043} & 0.029 & 0.121 & 0.048 & 0.173 & 0.051 \\
& 0.931 & 0.434 & \textbf{0.388} & 0.500 & \textbf{2.135} & 0.235 & \textbf{2.368} & 0.582 & 0.860 & \textbf{0.828} \\
\midrule
\multirow{3}{*}{DeepGPS} & 0.103 & \textbf{0.048} & 0.056 & \textbf{0.041} & 0.217 & \textbf{0.022} & \textbf{0.289} & \textbf{0.044} & \textbf{0.095} & 0.101\\
& 0.053 & \textbf{0.026} & 0.036 & \textbf{0.023} & 0.044 & \textbf{0.016} & \textbf{0.090} & \textbf{0.034} & \textbf{0.070} & \textbf{0.045}\\
& \textbf{0.726} & \textbf{0.393} & 0.470 & \textbf{0.290} & 2.599 & \textbf{0.103} & 2.573 & \textbf{0.440} & \textbf{0.576} & 0.870\\
\bottomrule
\end{tabular}
\begin{tablenotes}
      \small
      \item For each method, the top/middle/bottom row shows the root-mean-square/ median/max error. All errors are measured in meter. Note that the overall result is the average over all 30 scenes (see appendix), not only the scenes in this table.
    \end{tablenotes}
\centering
\end{table*}

\begin{figure*}
\includegraphics[trim=0 0 7mm 0,width=0.94\textwidth]{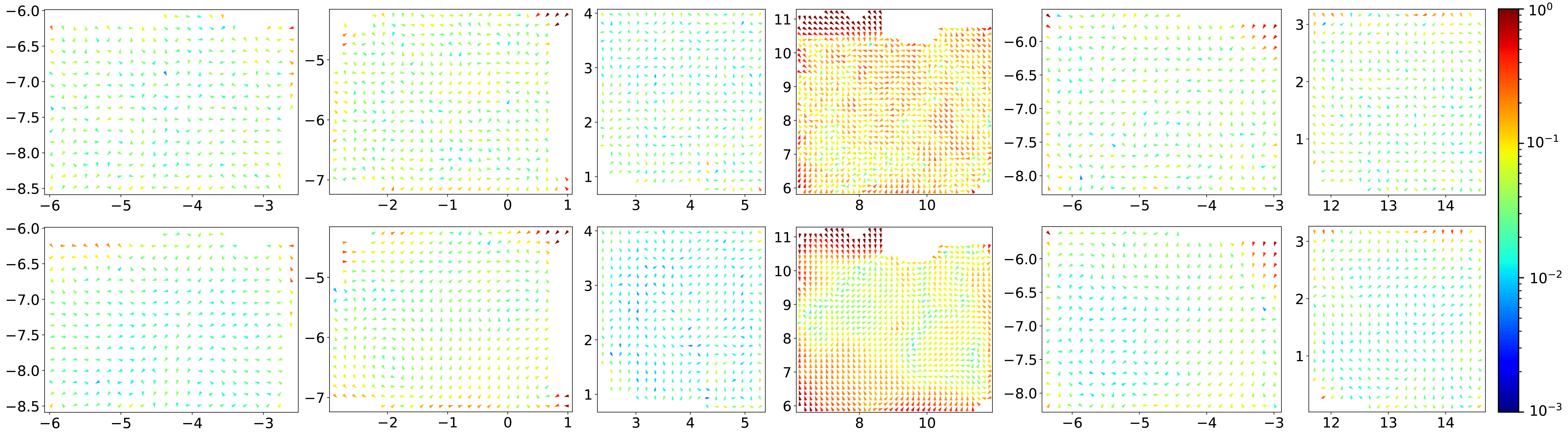}
\caption{\textbf{Sample visual results of positioning error in the virtualized photo-realistic environment experiments.} Top: PoseNet~\cite{Kendall:PoseNet:ICCV2015}. We show the localization errors computed from a discrete grid. The arrows indicate the error directions between the predicted positions and the ground truth, where the error magnitudes are color-coded.}
\vspace*{-4mm}
\label{fig:habitat-sim}
\end{figure*}

\subsection{Virtualized photo-realistic environment experiments}
\label{sec:photo-realistic}

\textbf{Data generation:}
We randomly picked 30 rooms in the Gibson~\cite{xiazamirhe2018gibsonenv} dataset. For each room, we collected about 8,000 images for training using the dense-sampling strategy 
and the testing images were collected on grids with 0.08m intervals in unoccupied areas. On each testing position, we placed the robot torwards 4 different orientations: $+x, -x, +y, -y$. The maximum error among the 4 orientations is used as the ATE at that location. The appendix shows sample original omnidirectional images taken in the experiment environment and the ones after the linear-polar warping (section~\ref{sec:trans_inv}).

    
    

\textbf{Training details:}
We use a modified ResNet18~\cite{he2016deep} as $f_{\theta}$ (section~\ref{sec:trans_inv}). The total number of epochs is 1,500 with a batch size 100. We apply an Adam optimizer with learning rate 0.0001.

\textbf{Results:}
Table~\ref{tab:cnn_sim} and Figure~\ref{fig:habitat-sim} show our quantitative and qualitative experiment results on selected scenes. Complete results are reported in the appendix. Our method shows comparable performance with PoseNet. However, the ground truth is more convenient to collect for our method. The relative error of our method is about 2\% (relative to the shorter side of the bounding box of the experiment environment). The main reason that our method has larger errors on some environments (e.g. the fourth column in Figure~\ref{fig:habitat-sim}, Islandton) is that our random-walk strategy may not be able to cover enough area, which also influences PoseNet. It is reasonable to expect a better performance by a better sampling strategy. Note that although the experiments are done in simulated environments, the results are still valuable since the images of scenes are photo-realistic.

\subsection{Real-world environment experiments}
\begin{table}[h!]
    \centering
    \caption{ATE for real-world experiments \label{tab:label_real}}
    \resizebox{0.45\textwidth}{!}{%
    \begin{tabular}{c|c c | c c | c c}
    \toprule
    \multirow{2}{*}{ATE (\si{m})} &  \multicolumn{2}{c}{RMS} & \multicolumn{2}{c}{Median} & \multicolumn{2}{c}{Max} \\
    \cline{2-7}
    & Mean & Std. & Mean & Std. & Mean & Std. \\ 
    \cline{1-7}
     PoseNet\cite{Kendall:PoseNet:ICCV2015} & 0.105& 0.053 & \textbf{0.032} & 0.005 & 0.762 & 0.457\\
    \cline{1-7}
     DeepGPS   & \textbf{0.053} & 0.021 & 0.036 & 0.004 & \textbf{0.175} & 0.026\\
    \bottomrule
    \end{tabular}
    }
    \begin{tablenotes}
      \small
      \item All errors are measured in meter.
    \end{tablenotes}
   \end{table}

\textbf{Data generation:}
We mounted two cameras on a TurtleBot\footnote{https://www.turtlebot.com/}: an omnidirectional camera that looks upwards and a perspective camera that looks horizontally. The images from the omnidirectional camera is used for providing visual localization input and the images from the perspective camera is used for obtaining the ground truth positions for evaluation. 
We placed 82 AprilTags~\cite{olson2011apriltag} in the environment. The 6-D poses of the tags were reconstructed using MASFM~\cite{feng2016marker}. Then the groud truth positions were obtained by the perspective camera via solving the PnP problem.
We collected 6 datasets in the environment using the end-point strategy and the train-test split for each dataset is 1,570/314. 

\textbf{Training details:} The network architecture and training setting is the same as we use in section~\ref{sec:photo-realistic}.

\begin{figure}[!t]
    \centering
    \includegraphics[width=0.33\textwidth]{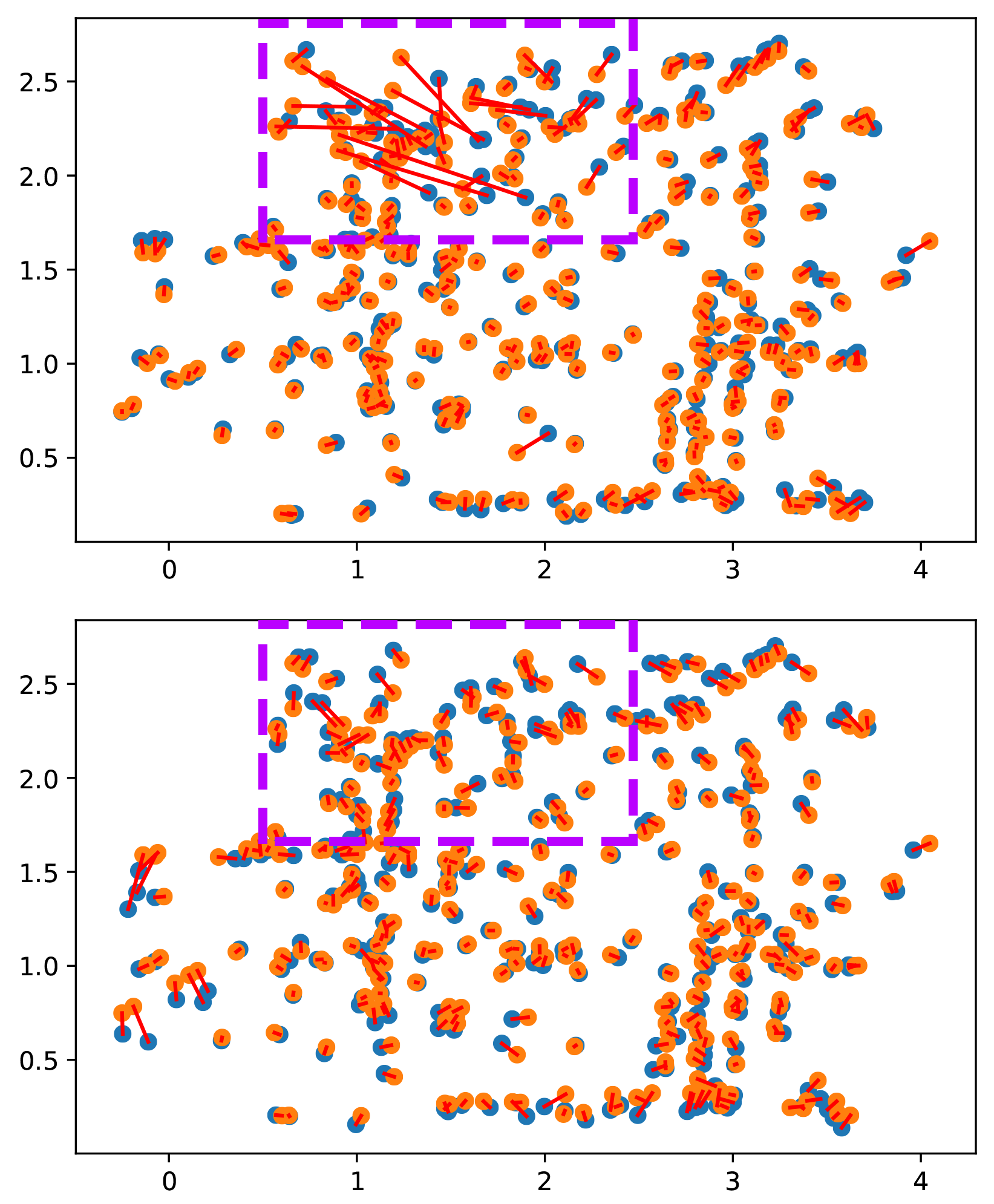}
    \caption{\textbf{Visual results of positioning error in the real-world environment experiments.} Top: PoseNet~\cite{Kendall:PoseNet:ICCV2015}, Bottom: DeepGPS. Blue dots are predicted positions and orange dots are ground truth positions. Red lines indicate the errors between them. The purple box shows the area where PoseNet has larger errors.}
    \label{fig:realworld_results}
    \vspace*{-6mm}
\end{figure}

\textbf{Results:}
Table~\ref{tab:label_real} lists the quantitative result of our method in the real-world environment. The mean and standard deviation of the ATEs are calculated among the 6 experiments. Interestingly, we found that the maximum ATE of our method is significantly smaller than PoseNet~\cite{Kendall:PoseNet:ICCV2015}, while the median ATE is very close between the two methods. Figure~\ref{fig:realworld_results} is one example of our experiments, which shows DeepGPS has fewer large errors than PoseNet. We believe it is because PoseNet suffers more from overfitting the positioning errors in the real-world ground truth. 


\section{Robustness Analysis}

We perform robustness analysis on our method to obtain better insights on it. Here, we study how the level of noise in the distance measurement and the number of training samples influence the positioning accuracy. We evaluate the influence of these two factors using both the toy example environment that has observations with correspondences and the virtualized photo-realistic environment.

\subsection{Level of noise}
We add random noise to each distance measurement between two consecutive positions. The noise-contaminated distance measurement $\Tilde{c}_{i,i+1}$ between two consecutive positions $\mathbf{p}_i$ and $\mathbf{p}_{i+1}$  can be modeled as:
\begin{equation}
\Tilde{c}_{i,i+1} =\Tilde{c}_{i,i+1} + n_{i,i+1}, 
\end{equation}
where $\Tilde{c}_{i,i+1}$ is the noise-free distance measurement, and $n_{i,i+1}$ is the random Gaussian noise $\mathcal{N}(0,\sigma_{i,i+1}^2)$, where the standard deviation $\sigma_{i,i+1}$ is proportional to noise-free distance, controlled by a factor $w$ i.e., $\sigma_{i,i+1}=wc_{i,i+1}$. In this experiment, we evaluate under $w = 0, 0.02, 0.04, 0.08, 0.10.$

\begin{figure}
\vspace{2mm}
    \centering
    \includegraphics[trim=5mm 5mm 0 0, width=0.3\textwidth]{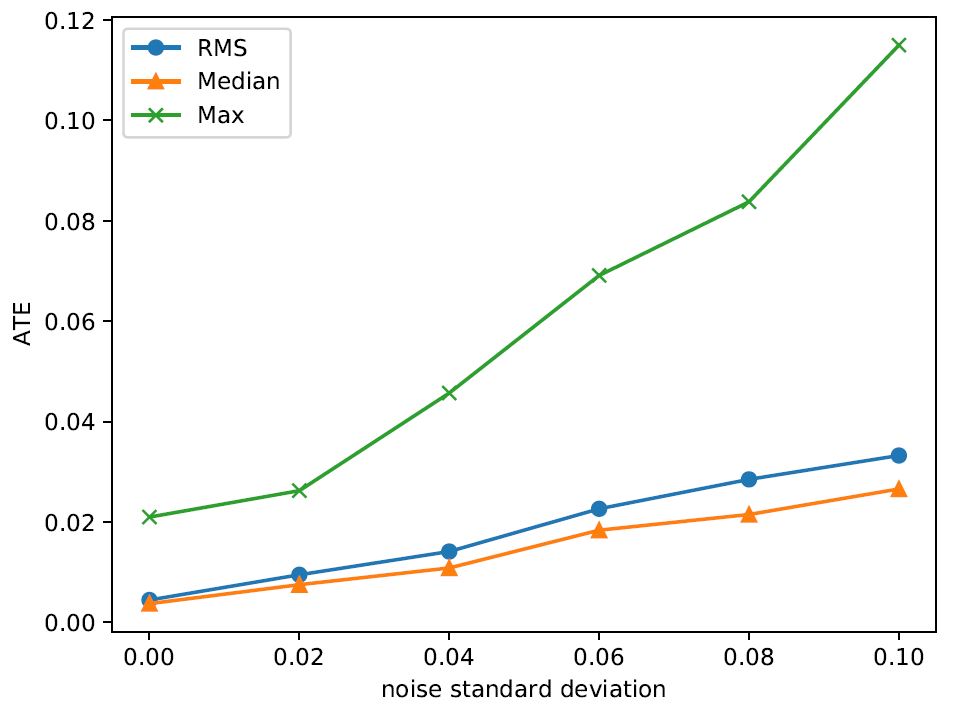} 
    \caption{\textbf{DeepGPS ATE vs. level of noise.}}
    \vspace*{-0.15in}
    \label{fig:ab_lon}
\end{figure}

Figure~\ref{fig:ab_lon} depicts the results. Although the max error grows faster with the increasing level of noise, the RMS and median are all less that 0.04 (2\%  relative to the side length of experiment environment), even if at 10\% of noise level. The results indicate that only few testing instances are sensitive to the noise. 
In practice, the noise level of a wheel encoder is less than 0.1\%, which is far less than the evaluation range (2.0\%-10.0\%). For example, the resolution of the wheel encoder of DYNAMIXEL-X series motor is $4096$ pulse/rev\footnote[1]{https://emanual.robotis.com/docs/en/dxl/x//\#dynamixel-x/}, which is equivalent to 0.02\% relative error for distance measurement. Therefore, our method is robust to the noise in distance measurement.

\subsection{Number of training samples}
\begin{figure}[!h]
\vspace{-2mm}
    \centering
    \includegraphics[trim=0 0 0 5mm, width=0.4\textwidth]{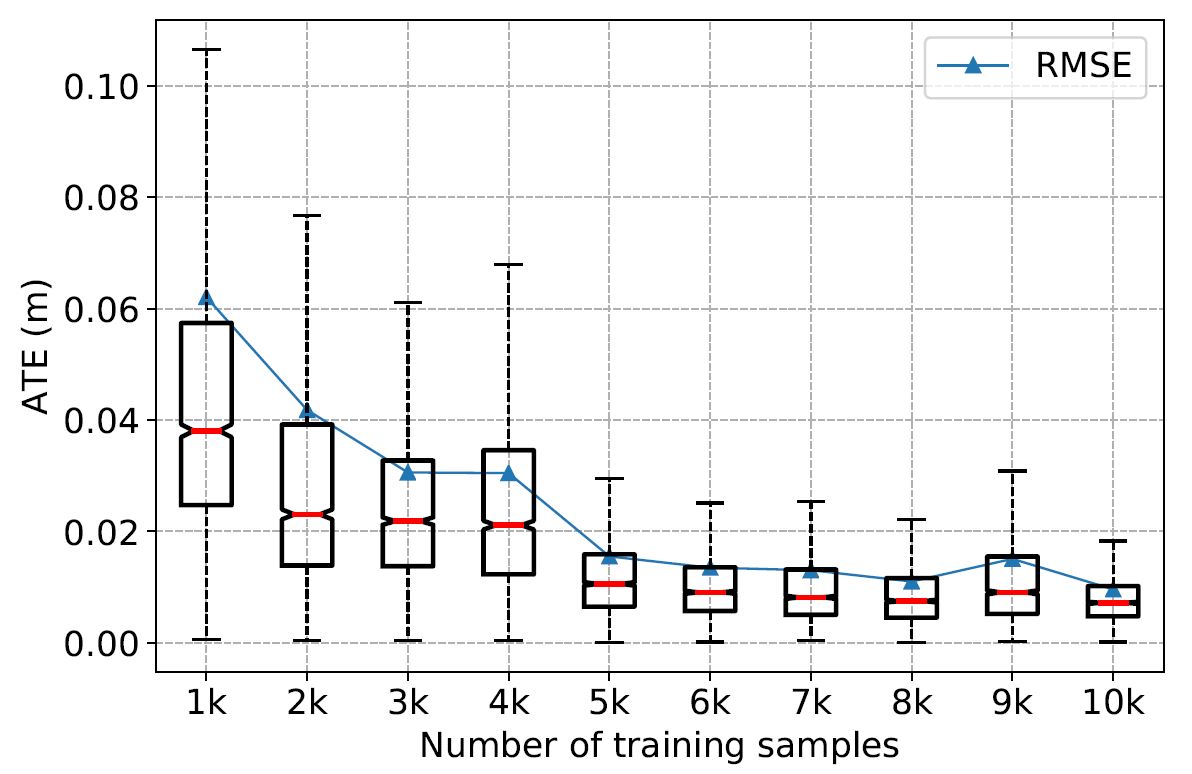}
    \caption{\textbf{DeepGPS ATE vs. number of training samples.}}
    \label{fig:ab_n}
    \vspace*{-0.15in}
\end{figure}

Figure~\ref{fig:ab_n} shows performance under different number of training samples. The dimension of the room is $7\,\si{m}\times4.3\,\si{m}$. Our method can achieve $0.04\,\si{m}$ accuracy with small number of training samples (1,000 samples, 33.2 samples/\si{m^2}). We notice that the ATE has a significant drop between $4,000$ to $5,000$ training samples, and after that the ATE almost remains at the same level. Therefore, for this living room, there is an optimal number of training samples, which is around 5,000. To increase the efficiency of the data collection, one of our future work is  to make our method training on-the-fly, i.e., the training and data collection process happen at the same time, and the data collection terminates when optimal number of training sample is reached. 
\section{Conclusion}



In this paper, we propose a new weakly-supervised positioning system: DeepGPS. This method uses distances between positions as supervision signal instead of directly using positions. In this way, the proposed method can significantly decrease the cost and difficulty of ground truth data collection while still maintaining high accuracy. The proposed method does not heavily depend on the modality of sensors, thus can be generalized to many conditions and environments. Besides, the proposed method is fully end-to-end, without explicitly models of the sensor and the environment, which saves the setup time and avoids data association, which is a non-trivial problem in traditional visual localization. One limitation of this method is that the performance will drop when the layout of the environment is highly concave. This problem can be alleviated by introducing sensors that can measure distances through obstacles.
In the future, we will extend our method to 3D case and apply it on large-scale environment.

\section*{Acknowledgment}
The research is supported by NSF CPS program under CMMI-1932187.


\small{
\bibliographystyle{IEEEtran}
\bibliography{ai4ce-tpl, Other, PlaceRecog, PositionSystem, VisualOdometry, WeaklyLearning}

\begin{thebibliography}{10}
\providecommand{\url}[1]{#1}
\csname url@rmstyle\endcsname
\providecommand{\newblock}{\relax}
\providecommand{\bibinfo}[2]{#2}
\providecommand\BIBentrySTDinterwordspacing{\spaceskip=0pt\relax}
\providecommand\BIBentryALTinterwordstretchfactor{4}
\providecommand\BIBentryALTinterwordspacing{\spaceskip=\fontdimen2\font plus
\BIBentryALTinterwordstretchfactor\fontdimen3\font minus
  \fontdimen4\font\relax}
\providecommand\BIBforeignlanguage[2]{{%
\expandafter\ifx\csname l@#1\endcsname\relax
\typeout{** WARNING: IEEEtran.bst: No hyphenation pattern has been}%
\typeout{** loaded for the language `#1'. Using the pattern for}%
\typeout{** the default language instead.}%
\else
\language=\csname l@#1\endcsname
\fi
#2}}

\bibitem{Newcombe:DTAM:ICCV11}
R.~A. {Newcombe}, S.~J. {Lovegrove}, and A.~J. {Davison}, ``{DTAM}: Dense
  tracking and mapping in real-time,'' in \emph{Proc. IEEE Int'l Conf. Computer
  Vision (ICCV)}, Nov 2011, pp. 2320--2327.

\bibitem{Tanskanen:LiveMetricReconMobile:ICCV13}
P.~{Tanskanen}, K.~{Kolev}, L.~{Meier}, F.~{Camposeco}, O.~{Saurer}, and
  M.~{Pollefeys}, ``Live metric {3D} reconstruction on mobile phones,'' in
  \emph{Proc. IEEE Int'l Conf. Computer Vision (ICCV)}, Dec 2013, pp. 65--72.

\bibitem{Lim:RTMono6DOF:IJRR15}
H.~Lim, S.~Sinha, M.~Cohen, M.~Uyttendaele, and H.~J. Kim, ``Real-time
  monocular image-based {6-DoF} localization,'' \emph{Int'l J. Robotics
  Research}, vol.~34, no. 4-5, pp. 476--492, April 2015.

\bibitem{Donoser:DiscrimFeat2PtMatch:CVPR14}
M.~Donoser and D.~Schmalstieg, ``Discriminative feature-to-point matching in
  image-based localization,'' in \emph{Proc. IEEE Conf. Comp.\ Vision and
  Pattern Recog. (CVPR)}, 2014, pp. 516--523.

\bibitem{Liu:Glb2D3DMatch:ICCV17}
L.~Liu, H.~Li, and Y.~Dai, ``Efficient global {2D-3D} matching for camera
  localization in a large-scale {3D} map,'' in \emph{Proc. IEEE Int'l Conf.
  Computer Vision (ICCV)}, 2017, pp. 2372--2381.

\bibitem{Ding:FuseSfMLidar:ICASSP17}
L.~Ding and G.~Sharma, ``Fusing structure from motion and lidar for dense
  accurate depth map estimation,'' in \emph{Proc. IEEE Int'l Conf. Acoustics,
  Speech, and Signal Processing (ICASSP)}, 2017, pp. 1283--1287.

\bibitem{taguchi2013point}
Y.~Taguchi, Y.-D. Jian, S.~Ramalingam, and C.~Feng, ``Point-plane slam for
  hand-held 3{D} sensors,'' in \emph{Proc. IEEE Int'l Conf. Robotics and Auto.
  (ICRA)}, 2013, pp. 5182--5189.

\bibitem{pumarola2017pl}
A.~Pumarola, A.~Vakhitov, A.~Agudo, A.~Sanfeliu, and F.~Moreno-Noguer,
  ``{PL-SLAM}: Real-time monocular visual slam with points and lines,'' in
  \emph{2017 IEEE international conference on robotics and automation (ICRA)},
  2017, pp. 4503--4508.

\bibitem{Fischler:RANSAC:CACM81}
M.~A. Fischler and R.~C. Bolles, ``Random sample consensus: a paradigm for
  model fitting with applications to image analysis and automated
  cartography,'' \emph{Commun. ACM}, vol.~24, no.~6, pp. 381--395, 1981.

\bibitem{Sarlin:Coare2FineCamLoc:CVPR19}
P.-E. Sarlin, C.~Cadena, R.~Siegwart, and M.~Dymczyk, ``From coarse to fine:
  Robust hierarchical localization at large scale,'' in \emph{Proc. IEEE Conf.
  Comp.\ Vision and Pattern Recog. (CVPR)}, 2019, pp. 12\,716--12\,725.

\bibitem{Avraham:EMPNet:ICCV19}
G.~Avraham, Y.~Zuo, T.~Dharmasiri, and T.~Drummond, ``{EMPNet}: Neural
  localisation and mapping using embedded memory points,'' in \emph{Proc. IEEE
  Int'l Conf. Computer Vision (ICCV)}, 2019, pp. 8120--8129.

\bibitem{Dube:SegMap:RSS18}
R.~Dub{\`e}, A.~Cramariuc, J.~Dugas, Daniel~andNieto, R.~Siegwart, and
  C.~Cadena, ``{SegMap}: {3D} segment mapping using data-driven descriptors,''
  in \emph{Proc. Robotics: Science and Systems (RSS)}, 2018.

\bibitem{Wang:DeepPCO:IROS19}
W.~Wang, M.~R.~U. Saputra, P.~Zhao, P.~Gusmao, B.~Yang, C.~Chen, A.~Markham,
  and N.~Trigoni, ``{DeepPCO}: End-to-end point cloud odometry through deep
  parallel neural network,'' in \emph{Proc. IEEE/RSJ Int'l Conf. Intelligent
  Robots and Systems (IROS)}, 2019, to appear.

\bibitem{Kendall:PoseNet:ICCV2015}
A.~Kendall, M.~Grimes, and R.~Cipolla, ``{PoseNet}: A convolutional network for
  real-time 6-{DOF} camera relocalization,'' in \emph{Proc. IEEE Int'l Conf.
  Computer Vision (ICCV)}, 2015, pp. 2938--2946.

\bibitem{Sattler:LimitationOfAPR:CVPR19}
T.~Sattler, Q.~Zhou, M.~Pollefeys, and L.~Leal-Taixe, ``Understanding the
  limitations of {CNN}-based absolute camera pose regression,'' in \emph{Proc.
  IEEE Conf. Comp.\ Vision and Pattern Recog. (CVPR)}, June 2019.

\bibitem{xiazamirhe2018gibsonenv}
F.~Xia, A.~R.~Zamir, Z.~He, A.~Sax, J.~Malik, and S.~Savarese, ``Gibson {Env}:
  real-world perception for embodied agents,'' in \emph{Proc. IEEE Conf. Comp.\
  Vision and Pattern Recog. (CVPR)}.\hskip 1em plus 0.5em minus 0.4em\relax
  IEEE, 2018.

\bibitem{brena2017evolution}
R.~F. Brena, J.~P. Garc{\'\i}a-V{\'a}zquez, C.~E. Galv{\'a}n-Tejada,
  D.~Mu{\~n}oz-Rodriguez, C.~Vargas-Rosales, and J.~Fangmeyer, ``Evolution of
  indoor positioning technologies: A survey,'' \emph{Journal of Sensors}, vol.
  2017, 2017.

\bibitem{zafari2019survey}
F.~Zafari, A.~Gkelias, and K.~K. Leung, ``A survey of indoor localization
  systems and technologies,'' \emph{IEEE Communications Surveys \& Tutorials},
  vol.~21, no.~3, pp. 2568--2599, 2019.

\bibitem{Lowe:SIFT:IJCV04}
D.~G. Lowe, ``Distinctive image features from scale-invariant keypoints,''
  \emph{Int'l J. Computer Vision}, vol.~60, no.~2, pp. 91--110, 2004.

\bibitem{Rublee:ORB:ICCV11}
E.~Rublee, V.~Rabaud, K.~Konolige, and G.~Bradski, ``{ORB}: An efficient
  alternative to {SIFT} or {SURF},'' in \emph{Proc. IEEE Int'l Conf. Computer
  Vision (ICCV)}, 2011.

\bibitem{Dube:SegMatch:ICRA17}
R.~Dub{\`e}, D.~Dugas, E.~Stumm, J.~Nieto, R.~Siegwart, and C.~Cadena,
  ``{SegMatch}: Segment based place recognition in {3D} point clouds,'' in
  \emph{Proc. IEEE Int'l Conf. Robotics and Auto. (ICRA)}, 2017, pp.
  5266--5272.

\bibitem{Brachmann:LearnLessMore6DCamLoc:CVPR18}
E.~Brachmann and C.~Rother, ``Learning less is more - {6D} camera localization
  via {3D} surface regression,'' in \emph{Proc. IEEE Conf. Comp.\ Vision and
  Pattern Recog. (CVPR)}, 2018.

\bibitem{Massiceti:RFvsNNCamLoc:ICRA17}
D.~Massiceti, A.~Krull, E.~Brachmann, C.~Rother, and P.~H. Torr, ``Random
  forests versus neural networks—what's best for camera localization?'' in
  \emph{Proc. IEEE Int'l Conf. Robotics and Auto. (ICRA)}, 2017, pp.
  5118--5125.

\bibitem{Taira:InLoc:CVPR18}
H.~Taira, M.~Okutomi, T.~Sattler, M.~Cimpoi, M.~Pollefeys, J.~Sivic, T.~Pajdla,
  and A.~Torii, ``{InLoc}: Indoor visual localization with dense matching and
  view synthesis,'' in \emph{Proc. IEEE Conf. Comp.\ Vision and Pattern Recog.
  (CVPR)}, 2018.

\bibitem{Schonberger:SemanticVisLoc:CVPR18}
J.~L. Sch{\`o}nberger, M.~Pollefeys, A.~Geiger, and T.~Sattler, ``Semantic
  visual localization,'' in \emph{Proc. IEEE Conf. Comp.\ Vision and Pattern
  Recog. (CVPR)}, 2018, pp. 6896--6906.

\bibitem{SzegedyGoogLeNet:CVPR15}
C.~Szegedy, W.~Liu, Y.~Jia, P.~Sermanet, S.~Reed, D.~Anguelov, D.~Erhan,
  V.~Vanhoucke, and A.~Rabinovich, ``Going deeper with convolutions,'' in
  \emph{Proc. IEEE Conf. Comp.\ Vision and Pattern Recog. (CVPR)}, 2015.

\bibitem{Kendall:GeoLoss4PoseReg:CVPR2017}
A.~Kendall and R.~Cipolla, ``Geometric loss functions for camera pose
  regression with deep learning,'' in \emph{Proc. IEEE Conf. Comp.\ Vision and
  Pattern Recog. (CVPR)}, July 2017.

\bibitem{Melekhov:CamLocHourglassNet:ICCVW17}
I.~Melekhov, J.~Ylioinas, J.~Kannala, and E.~Rahtu, ``Image-based localization
  using hourglass networks,'' in \emph{Proc. IEEE Int'l Conf. Computer Vision
  Wksp. (ICCV Wksp.)}, 2017, pp. 879--886.

\bibitem{Henriques:MapNetRNN:CVPR18}
J.~F. Henriques and A.~Vedaldi, ``Map{N}et: An allocentric spatial memory for
  mapping environments,'' in \emph{Proc. IEEE Int'l Conf. Computer Vision
  (ICCV)}, 2018.

\bibitem{Wu:DelveCNN4CamLoc:ICRA2017}
J.~Wu, L.~Ma, and X.~Hu, ``Delving deeper into convolutional neural networks
  for camera relocalization,'' in \emph{Proc. IEEE Int'l Conf. Robotics and
  Auto. (ICRA)}, 2017, pp. 5644--5651.

\bibitem{Naseer:RegMonoCam6DOFOutdoor:IROS2017}
T.~Naseer and W.~Burgard, ``Deep regression for monocular camera-based 6-{DoF}
  global localization in outdoor environments,'' in \emph{Proc. IEEE/RSJ Int'l
  Conf. Intelligent Robots and Systems (IROS)}, 2017, pp. 1525--1530.

\bibitem{Brahmbhatt:GeoLearnMap4CamLoc:CVPR18}
S.~Brahmbhatt, J.~Gu, K.~Kim, J.~Hays, and J.~Kautz, ``Geometry-aware learning
  of maps for camera localization,'' in \emph{Proc. IEEE Conf. Comp.\ Vision
  and Pattern Recog. (CVPR)}, June 2018.

\bibitem{Radwan:VLocNetPP:RAL18}
N.~{Radwan}, A.~{Valada}, and W.~{Burgard}, ``{VLocNet++}: Deep multitask
  learning for semantic visual localization and odometry,'' \emph{IEEE Robotics
  and Automation Letters}, vol.~3, no.~4, pp. 4407--4414, Oct 2018.

\bibitem{Fraundorfer:VOSurveyPart2:RAM12}
F.~{Fraundorfer} and D.~{Scaramuzza}, ``Visual odometry: Part {II}: Matching,
  robustness, optimization, and applications,'' \emph{IEEE Robotics Automation
  Mag.}, vol.~19, no.~2, pp. 78--90, June 2012.

\bibitem{Wang:DeepVO:ICRA2017}
S.~Wang, R.~Clark, H.~Wen, and N.~Trigoni, ``{DeepVO}: Towards end-to-end
  visual odometry with deep recurrent convolutional neural networks,'' in
  \emph{Proc. IEEE Int'l Conf. Robotics and Auto. (ICRA)}, 2017, pp.
  2043--2050.

\bibitem{Clark:VidLoc:CVPR2017}
R.~Clark, S.~Wang, A.~Markham, N.~Trigoni, and H.~Wen, ``{VidLoc}: A deep
  spatio-temporal model for 6-dof video-clip relocalization,'' in \emph{Proc.
  IEEE Conf. Comp.\ Vision and Pattern Recog. (CVPR)}, 2017, pp. 6856--6864.

\bibitem{Ma:SelSupLidar&cam:ICRA2019}
F.~Ma, G.~V. Cavalheiro, and S.~Karaman, ``Self-supervised sparse-to-dense:
  Self-supervised depth completion from lidar and monocular camera,'' in
  \emph{Proc. IEEE Int'l Conf. Robotics and Auto. (ICRA)}, 2019, pp.
  3288--3295.

\bibitem{Pillai:TowVEGO;IROS2017}
S.~Pillai and J.~J. Leonard, ``Towards visual ego-motion learning in robots,''
  in \emph{Proc. IEEE/RSJ Int'l Conf. Intelligent Robots and Systems (IROS)},
  2017, pp. 5533--5540.

\bibitem{Wang:LearnInertia:ICASSP2019}
C.~Wang, Y.~Yuan, and Q.~Wang, ``Learning by inertia: Self-supervised monocular
  visual odometry for road vehicles,'' in \emph{Proc. IEEE Int'l Conf.
  Acoustics, Speech, and Signal Processing (ICASSP)}, 2019, pp. 2252--2256.

\bibitem{Zhou:SfMLearner:CVPR17}
T.~Zhou, M.~Brown, N.~Snavely, and D.~G. Lowe, ``Unsupervised learning of depth
  and ego-motion from video,'' in \emph{Proc. IEEE Conf. Comp.\ Vision and
  Pattern Recog. (CVPR)}, 2017, pp. 1851--1858.

\bibitem{Li:UnDeepVO:ICRA2018}
R.~Li, S.~Wang, Z.~Long, and D.~Gu, ``{UnDeepVO}: Monocular visual odometry
  through unsupervised deep learning,'' in \emph{Proc. IEEE Int'l Conf.
  Robotics and Auto. (ICRA)}, 2018, pp. 7286--7291.

\bibitem{Ding:DeepMapping:CVPR19}
L.~Ding and C.~Feng, ``{DeepMapping}: Unsupervised map estimation from multiple
  point clouds,'' in \emph{Proc. IEEE Conf. Comp.\ Vision and Pattern Recog.
  (CVPR)}, June 2019.

\bibitem{Stewart:PhyConstraintSupervision:AAAI17}
R.~Stewart and S.~Ermon, ``Label-free supervision of neural networks with
  physics and domain knowledge,'' in \emph{Proc. the Thirty-First AAAI
  Conference on Artificial Intelligence (AAAI)}, 2017, pp. 2576--2582.

\bibitem{kayhan2020translation}
O.~S. Kayhan and J.~C.~v. Gemert, ``On translation invariance in cnns:
  Convolutional layers can exploit absolute spatial location,'' in \emph{Proc.
  IEEE Conf. Comp.\ Vision and Pattern Recog. (CVPR)}, 2020, pp.
  14\,274--14\,285.

\bibitem{he2016deep}
K.~He, X.~Zhang, S.~Ren, and J.~Sun, ``Deep residual learning for image
  recognition,'' in \emph{Proc. IEEE Conf. Comp.\ Vision and Pattern Recog.
  (CVPR)}, 2016, pp. 770--778.

\bibitem{sturm2012benchmark}
J.~Sturm, N.~Engelhard, F.~Endres, W.~Burgard, and D.~Cremers, ``A benchmark
  for the evaluation of {RGB-D SLAM} systems,'' in \emph{Proc. IEEE/RSJ Int'l
  Conf. Intelligent Robots and Systems (IROS)}, 2012, pp. 573--580.

\bibitem{Kingma:Adam:ICLR15}
D.~P. Kingma and J.~Ba, ``Adam: A method for stochastic optimization,'' in
  \emph{Int'l Conf. Learning Representations (ICLR)}, 2015.

\bibitem{olson2011apriltag}
E.~Olson, ``Apriltag: A robust and flexible visual fiducial system,'' in
  \emph{Proc. IEEE Int'l Conf. Robotics and Auto. (ICRA)}, 2011, pp.
  3400--3407.

\bibitem{feng2016marker}
C.~Feng, V.~R. Kamat, and C.~C. Menassa, ``Marker-assisted structure from
  motion for 3d environment modeling and object pose estimation,'' in
  \emph{Construction Research Congress 2016}, 2016, pp. 2604--2613.

\bibitem{habitat19iccv}
M.~Savva, A.~Kadian, O.~Maksymets, Y.~Zhao, E.~Wijmans, B.~Jain, J.~Straub,
  J.~Liu, V.~Koltun, J.~Malik, D.~Parikh, and D.~Batra, ``Habitat: {A}
  {P}latform for {E}mbodied {AI} {R}esearch,'' in \emph{Proc. IEEE Int'l Conf.
  Computer Vision (ICCV)}, 2019.

\end{thebibliography}
}

\onecolumn

\begin{multicols}{2}
\appendix
\section{Supplementary Material}

\subsection{More results for photo-realistic environment experiment}
Table~\ref{tab:cnn_sim_sup} shows more experimental results in the Habitat-sim~\cite{habitat19iccv} environment.

\subsection{Experiment with a simulated 2D environment using Lidar}

\textbf{Experimental settings.} We use the simulator in~\cite{Ding:DeepMapping:CVPR19} to create four 2D environments and a virtual 2D Lidar scanner scans point clouds in these environments. Each local scan has 256 points which are arranged in order. We choose to use an MLP as $f_\theta(\cdot)$. The number of neurons in each layer of $f_\theta(\cdot)$ is 512, 512, 512, 1024, 512, 512, 256, 256, 128, 2 respectively. The input layer is the flattened 2D coordinates of the point clouds. $N$ is set to 10,000, and $R$ is set to 100. In each training iteration, at each position, 5 observations are randomly selected from the 100 different orientations. The testing set used for quantitative evaluation has 2,000 samples. 

\textbf{Baseline method.} We compare our methods with K-Nearest Neighbor (KNN), which finds the closest instance in the training set to the query observation and use the ground truth position attached to that instance as the estimated position. However, it is unfeasible to apply KNN on the raw observation space, due to the high dimension. Therefore, we apply Principle Component Analysis (PCA) to reduce the dimension from 512 to 128 before KNN.

\begin{figure*}[t!]
    \centering
    \includegraphics[width=1\textwidth, trim={0 0 0  0}, clip]{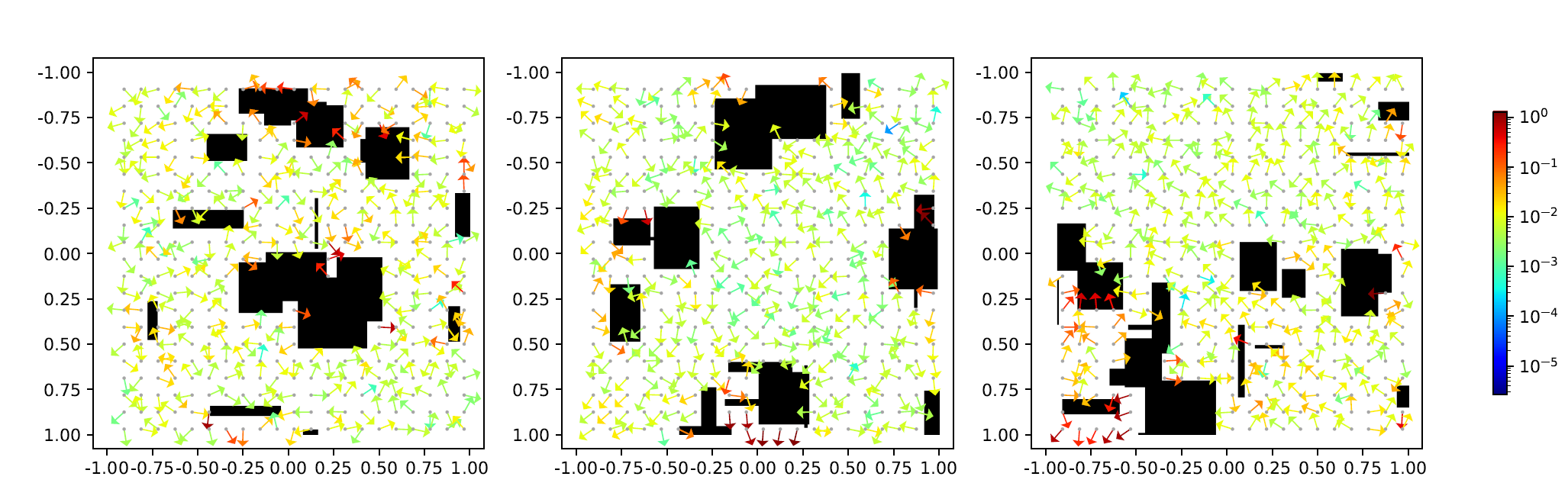}

    \caption{Visual results of localization error on the 2D point cloud datasets. Each image shows a unique environment where the black pixels are the occupied by obstacles. For a clear visualization, we show the localization errors computed from a discrete grid (gray dots). The arrows indicate the error directions between the predicted positions and the ground truth, where the error magnitudes are color-coded. Best viewed in color.}
    \label{fig:pcd_quali}
\end{figure*}

\textbf{Results.} On the 2D Lidar point cloud dataset. Our method has better performance on median error and PCA+KNN method has better performance on RMSE and Max error (see Table~\ref{tab:pcd_acc}). This is because the positions in our training set have very large density and the PCA+KNN method can almost memorize each location in the environment. However, memory efficiency is very low for PCA+KNN. Our method only takes about 2 million float number of memory during operation while PCA+KNN takes about 66 million, which is 33 times of our method. Bedsides, the memory usage grows with the increasing size of training set.  Again, our method does not need ground truth positions. Figure~\ref{fig:pcd_quali} illustrates the spatial distribution of the error.  We found that the area that are surrounded by obstacles has the largest error. This is because that our method relies on the distances between locations. The obstacles surrounding an area limit the measurement of distance from many directions. Therefore, the constraints on the estimated positions in such an area are not strong enough, which makes estimation degenerated. One way to improve this is to apply techniques that can measure distance through obstacles, for instance, Wi-Fi. In this case, we can have many robots in the same environment. The distances between robots can be measured through obstacles, which provide more constraints on the position estimation in addition to the distances measured by wheel encoder. 

\subsection{Example images collected in Habitat-sim~\cite{habitat19iccv}}
\begin{figure*}[h!]
    \centering
    \includegraphics[width=0.3\textwidth]{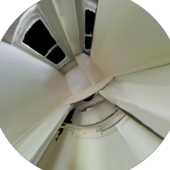}\includegraphics[width=0.3\textwidth]{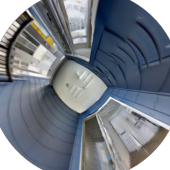}\includegraphics[width=0.3\textwidth]{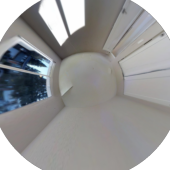} \\

     \includegraphics[angle=270,width=0.3\textwidth]{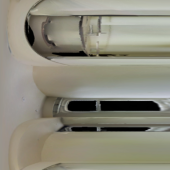}\includegraphics[angle=270, width=0.3\textwidth]{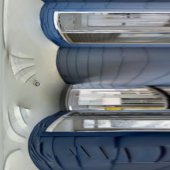}\includegraphics[angle=270,width=0.3\textwidth]{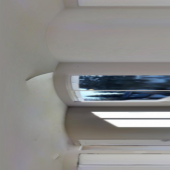}
    \caption{\textbf{Example images collected via Habitat-Sim~\cite{habitat19iccv}.} Top row: original images; Bottom row: images after linear-polar warping. From left to right: Delton, Montreal0, Pasatiempo0}
    \label{fig:sim_warp}
    \vspace{-5mm}
\end{figure*}

Figure~\ref{fig:sim_warp} shows some example images collected in the Habitat-sim~\cite{habitat19iccv} environment.
\end{multicols}

\begin{table*}[h!]
\centering
    \caption{ATE for experiment in Habitat-Sim.}
    \resizebox{1.0\textwidth}{!}{%
\begin{tabular}{
m{1.5cm} |  
>{\centering}m{0.8cm} 
>{\centering}m{1.3cm} 
>{\centering}m{1cm} 
>{\centering}m{1.3cm}
>{\centering}m{1.3cm}
>{\centering}m{1.3cm} 
>{\centering}m{1.3cm}
>{\centering}m{1.3cm}
>{\centering}m{1cm} 
>{\centering\arraybackslash}m{1.5cm} 
}
\toprule
ATE (m) & Aloha1  & Arona & Arona1 & Avonia0 & Avonia1 & Barranquitas0 & Barranquitas1 & Barranquitas2 & Cokeville1 & Eagerville
\\
\midrule
\multirow{3}{*}{PoseNet~\cite{Kendall:PoseNet:ICCV2015}} & 0.119 & 0.063 & \textbf{0.052} & \textbf{0.071} & 0.128 & \textbf{0.114} & 0.101 & \textbf{0.047} & 0.049 & \textbf{0.060}\\ 
& 0.049 & 0.037 & 0.042 & \textbf{0.043} & 0.045 & \textbf{0.088} & 0.081 & 0.034 & 0.031 & 0.042\\
& \textbf{1.514} & 1.054 & \textbf{0.448} & \textbf{0.881} & 1.428 & 0.945 & 0.554 & 0.408 & 0.401 & \textbf{0.680} \\
\midrule
\multirow{3}{*}{DeepGPS} & \textbf{0.103} & \textbf{0.031} & 0.079 & 0.187 & \textbf{0.095} & 0.165 & \textbf{0.062} & 0.051 & \textbf{0.034} & 0.069\\
& \textbf{0.027} & \textbf{0.019} & \textbf{0.034} & 0.073 & \textbf{0.030} & 0.103 & \textbf{0.032} & \textbf{0.026} & \textbf{0.019} & \textbf{0.031}\\
& 1.741 & \textbf{0.409} & 0.850 & 1.055 & \textbf{1.350} & \textbf{0.695} & \textbf{0.524} & \textbf{0.334} & \textbf{0.294} & 0.997\\
\bottomrule

\toprule
ATE (m)  & Hercules & Kemblesville & Montreal1 & Pasatiempo0 & Pasatiempo1 & Rancocas0 & Rancocas1 & Roxboro0 & Sawpit0 & Sodaville0 
\\
\midrule
\multirow{3}{*}{PoseNet~\cite{Kendall:PoseNet:ICCV2015}} & 0.083 & 0.090 & \textbf{0.118} & \textbf{0.048} & \textbf{0.036} & 0.050 & 0.049 & 0.077 & 0.061 & \textbf{0.115}\\ 
& 0.031 & 0.039 & \textbf{0.054} & 0.037 & 0.030 & 0.041 & 0.039 & 0.049 & 0.045 & \textbf{0.069}\\
& 1.140 & 1.122 & \textbf{1.356} & \textbf{0.260} & \textbf{0.222} & 0.319 & \textbf{0.291} & 0.686 & 0.332 & \textbf{1.277}\\
\midrule
\multirow{3}{*}{DeepGPS} & \textbf{0.066} & \textbf{0.088} & 0.439 & \textbf{0.048} & 0.068 & \textbf{0.034} & \textbf{0.045} & \textbf{0.045} & \textbf{0.052} & 0.272\\
& \textbf{0.024} & \textbf{0.030} & 0.217 & \textbf{0.026} & \textbf{0.024} & \textbf{0.030} & \textbf{0.029} & \textbf{0.034} & \textbf{0.033} & 0.071\\
& \textbf{1.052} & \textbf{0.867} & 1.984 & 0.365 & 1.230 & \textbf{0.259} & 0.359 & \textbf{0.376} & \textbf{0.330} & 2.002\\
\bottomrule

\toprule
ATE (m)  & Wells &  Aloha  & Brevort & Cokeville & Delton  & Euharlee & Germfask & Islandton & Montreal & Sodaville  
\\
\midrule
\multirow{3}{*}{PoseNet~\cite{Kendall:PoseNet:ICCV2015}} & 0.115 & \textbf{0.061} & 0.060 & \textbf{0.040} & 0.060 & \textbf{0.161} & 0.037 & 0.338 & 0.072 & 0.211 \\ 
& 0.040 & \textbf{0.037} & 0.043 & \textbf{0.032} & 0.035 & \textbf{0.043} & 0.029 & 0.121 & 0.048 & 0.173 \\
& 1.085 & 0.931 & 0.434 & \textbf{0.388} & 0.500 & \textbf{2.135} & 0.235 & \textbf{2.368} & 0.582 & 0.860 \\
\midrule
\multirow{3}{*}{DeepGPS} & \textbf{0.088} & 0.103 & \textbf{0.048} & 0.056 & \textbf{0.041} & 0.217 & \textbf{0.022} & \textbf{0.289} & \textbf{0.044} & \textbf{0.095} \\ 
& \textbf{0.033} &0.053 & \textbf{0.026} & 0.036 & \textbf{0.023} & 0.044 & \textbf{0.016} & \textbf{0.090} & \textbf{0.034} & \textbf{0.070} \\
& \textbf{0.843} & \textbf{0.726} & \textbf{0.393} & 0.470 & \textbf{0.290} & 2.599 & \textbf{0.103} & 2.573 & \textbf{0.440} & \textbf{0.576} \\
\bottomrule

\end{tabular}
}
\begin{tablenotes}
      \small
      \item For each method, the top/middle/bottom row shows the root-mean-square/median/max error. All errors are measured in meter.
    \end{tablenotes}

\centering
\label{tab:cnn_sim_sup} 
\end{table*}

\begin{table}[h!]
\centering
    \caption{ATE for 2D Lidar point cloud. }
\centering
\resizebox{0.5\textwidth}{!}{%
\begin{tabular}{>{\centering\arraybackslash}m{0.8cm} >{\centering\arraybackslash}m{0.5cm} >{\centering\arraybackslash}m{0.5cm} >{\centering\arraybackslash}m{0.5cm} | >{\centering\arraybackslash}m{0.5cm} >{\centering\arraybackslash}m{0.5cm} >{\centering\arraybackslash}m{0.5cm}}
\toprule
\multirow{2}{*}{ATE} & \multicolumn{3}{c}{DeepGPS} & \multicolumn{3}{c}{PCA+KNN} \\ 
\cline{2-7}
 & RMSE & Median & Max & RMSE & Median & Max \\ \midrule
\textbf{Env. 1} & 0.162 & \textbf{0.008} & 2.449 & \textbf{0.142} & 0.013 & \textbf{1.980} \\ 
 \textbf{Env. 2} & 0.177 & \textbf{0.006} & 2.146 & \textbf{0.151} & 0.014 & \textbf{1.981}\\
\textbf{Env. 3} & 0.178 & \textbf{0.008} & 2.545 & \textbf{0.177} & 0.015 & \textbf{2.346} \\ 
\textbf{Env. 4} & 0.164 & \textbf{0.008} & 2.510 & \textbf{0.150} & 0.014 & \textbf{2.398}\\
\bottomrule
\end{tabular}}
\label{tab:pcd_acc}
\end{table}

\end{document}